%% file: main.tex
\RequirePackage{fix-cm}
\documentclass[twocolumn]{svjour3}          
\smartqed  
\usepackage{graphicx}
\usepackage{siunitx} 
\usepackage[center]{caption} 
\usepackage{bm} 
\usepackage{soul} 
\usepackage{xcolor}
\usepackage{booktabs}
\usepackage{amsmath,amsfonts,amssymb}
\usepackage{algorithmic}
\usepackage{algorithm}
\usepackage{array}
\usepackage{textcomp}
\usepackage{stfloats}
\usepackage{url}
\usepackage{verbatim}
\usepackage{graphicx}
\usepackage{cite}
\usepackage{caption}
\usepackage{xcolor}
\usepackage{soul}
\usepackage[printonlyused]{acronym}
\usepackage{comment}
\usepackage{multirow}
\usepackage{amssymb}
\usepackage{bm}
\usepackage{nicefrac}
\usepackage{tikz}
\usepackage[utf8]{inputenc}
\usepackage{multirow}
\usepackage{rotating}
\usepackage{cuted}
\usepackage{cite}
\usepackage{flushend}
\usepackage{lineno}
\usepackage{subfig}
\usepackage[hidelinks]{hyperref}
\usepackage{float}
\restylefloat{table}
\usepackage{siunitx}

\definecolor{blue}{RGB}{41,5,195} 
\definecolor{red}{RGB}{255,0,0} 
\definecolor{skyblue}{RGB}{135, 206, 235} 

\soulregister\ref7
\soulregister\cite7
\soulregister\ac7
\soulregister\acl7
\soulregister\acs7
\soulregister\acf7
\soulregister\acp7
\soulregister\iac7

\newcounter{issuecntr}

\newcommand{\reffig}[1]{Fig. \ref{#1}}
\newcommand{\reftab}[1]{Table \ref{#1}}

\newcommand{\refasm}[1]{Assumption \ref{#1}}

\newcommand{\real}{\mathbb{R}}
\newcommand{\minus}{\scalebox{0.75}[1.0]{$-$}}

\newtheorem{assumption}{Assumption}

\usepackage{tikz}
\usepackage{program}
\usepackage{pgfplots}
\pgfplotsset{compat=1.14}
\usetikzlibrary{backgrounds,arrows,automata,shapes,positioning,calc,through,spy}
\pgfdeclarelayer{background}
\pgfdeclarelayer{foreground}
\pgfsetlayers{background,main,foreground}

\input{tikz_config}

\tikzset{new spy style/.style={spy scope={%
  magnification=5,
  size=1.25cm,
  connect spies,
  every spy on node/.style={
    rectangle,
    draw,
  },
  every spy in node/.style={
    draw,
    rectangle,
    fill=white
  }
  }
  }
}

\newcommand{\PREPRINTYEAR}{2024}
\newcommand{\PUBLISHEDIN}{Journal of Intelligent and Robotics Systems}
\newcommand{\DOI}{10.1007/s10846-024-02108-0} 

\usepackage[placement=top,vshift=-7]{background}
\SetBgScale{1.0}
\SetBgContents{\PUBLISHEDIN. PREPRINT VERSION - DO NOT DISTRIBUTE. \href{https://doi.org/\DOI}{DOI \DOI}}
\SetBgColor{black}
\SetBgAngle{0}
\SetBgOpacity{1.0}

\begin{document}

\thispagestyle{empty}
\onecolumn
{
  \topskip0pt
  \vspace*{\fill}
  \centering
  \LARGE{%
    \copyright{} \PREPRINTYEAR~\PUBLISHEDIN\\
    \vspace{1cm}
    Personal use of this material is permitted.
    Permission from \PUBLISHEDIN~must be obtained for all other uses, in any current or future media, including reprinting or republishing this material for advertising or promotional purposes, creating new collective works, for resale or redistribution to servers or lists, or reuse of any copyrighted component of this work in other works.}
    \vspace*{\fill}
}
\NoBgThispage
\twocolumn          	
\BgThispage

\input{definitions}

\title{A Minimalistic 3D Self-Organized UAV Flocking Approach for Desert Exploration}


\author{Thulio Amorim \and
        Tiago Nascimento \and
        Akash Chaudhary \and
        Eliseo Ferrante \and
        Martin Saska
}


\institute{T. Amorim and T. Nascimento \at
              Department of Computer Systems \\
              Universidade Federal da Paraíba\\
              Brazil \\
              \email{laser@ci.ufpb.br}
           \and
           A. Chaudhary, T. Nascimento and M. Saska \at
              Cybernetics Department \\
              Czech Technical University in Prague \\
              Czech Republic \\
              http://mrs.felk.cvut.cz/
            \and
            E. Ferrante \at
            Technology Innovation Institute \\
            United Arab Emirates
}

\date{Received: date / Accepted: date}

\maketitle

\begin{abstract}
In this work, we propose a minimalistic swarm flocking approach for multirotor unmanned aerial vehicles (UAVs). Our approach allows the swarm to achieve cohesively and aligned flocking (collective motion), in a random direction, without externally provided directional information exchange (alignment control). The method relies on minimalistic sensory requirements as it uses only the relative range and bearing of swarm agents in local proximity obtained through onboard sensors on the UAV. Thus, our method is able to stabilize and control the flock of a general shape above a steep terrain without any explicit communication between swarm members. To implement proximal control in a three-dimensional manner, the Lennard-Jones potential function is used to maintain cohesiveness and avoid collisions between robots. The performance of the proposed approach was tested in real-world conditions by experiments with a team of nine UAVs. Experiments also present the usage of our approach on UAVs that are independent of external positioning systems such as the Global Navigation Satellite System (GNSS). Relying only on a relative visual localization through the ultraviolet direction and ranging (UVDAR) system, previously proposed by our group, the experiments verify that our system can be applied in GNSS-denied environments. The degree achieved of alignment and cohesiveness was evaluated using the metrics of order and steady-state value.
\end{abstract}

\section{Introduction}
\label{sec:introduction}

\begin{figure}[t]
    \centering
    \includegraphics[width=0.5\textwidth]{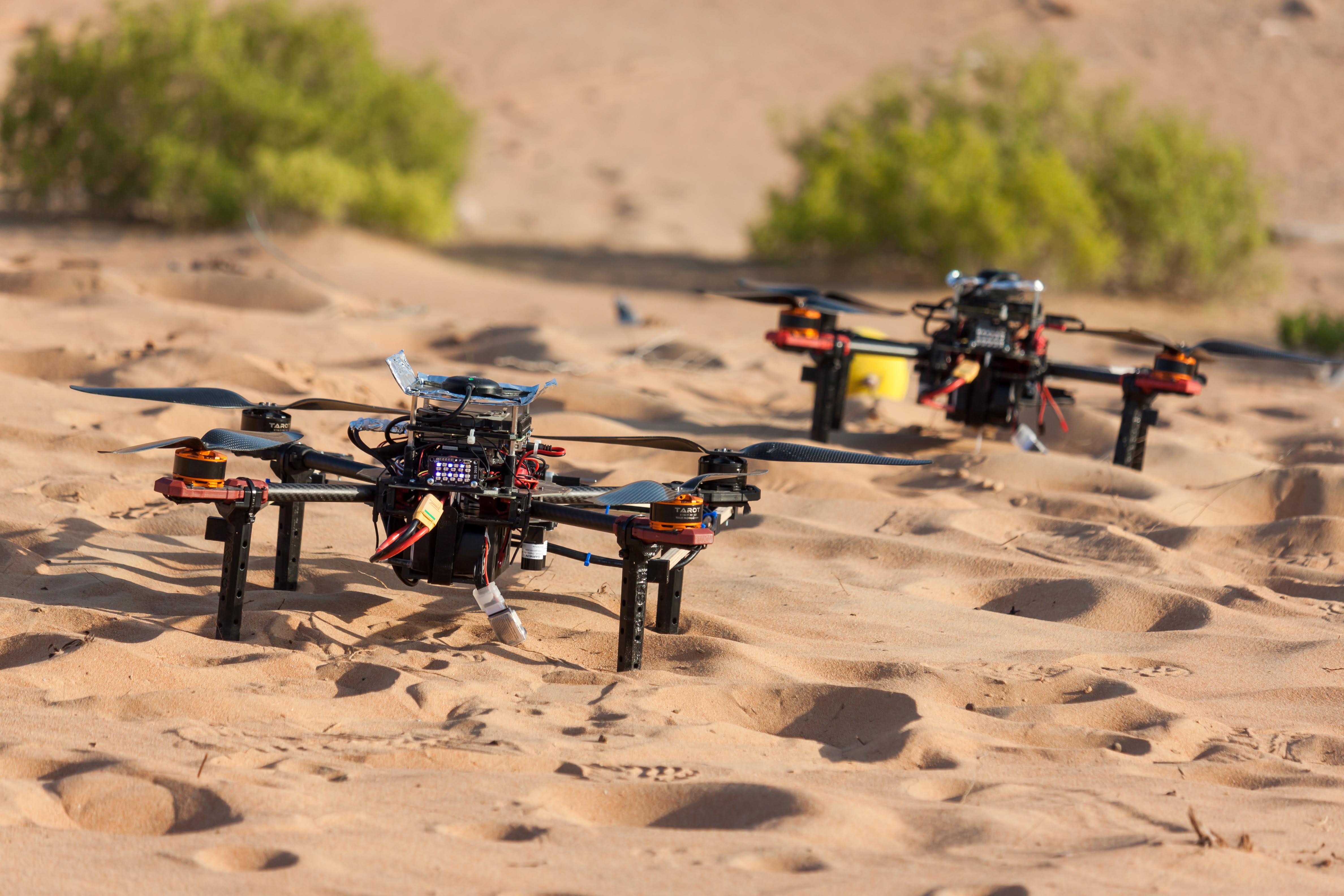}
    \caption{Swarm of unmanned aerial vehicles in the desert using our robotic platform.}
    \label{fig:swarms}
\end{figure}

Flocking is a coordinated motion of a large group of individuals moving together towards the same target direction (see \reffig{fig:swarms}). This collective behavior can be observed in different species in nature and emulating it in artificial systems has been an active research topic inspired by the fact of how easily simple individuals with limited resources can achieve complex organizations. In robotics, different methods have been proposed to accomplish the flocking behavior in multi-robot systems. Those works can be distinguished based on particular properties, such as the specification of the robot used as an individual and the environment in which the system can be applied. For example, numerous methods have already been proposed for ground robots acting in a plain environment with external global localization \cite{Ferrante:2012}.

To accomplish the flocking behavior in the \ac{3D} space using flying robots, designing methods with the same desirable attributes, such as scalability, robustness, and flexibility, is still an open problem. Add to that also the need to cope with the limitations on state estimation, such as no use of global information that would restrict the operational space of designed swarming systems. In most state-of-the-art solutions, robot flocking is performed by combining two main control techniques: proximal control and alignment control \cite{Brambilla:2013}. The proximal control uses the range and bearing of the neighboring agents to maintain the distance needed to achieve a safe flocking behavior. There is no standard sensor available to obtain the range and bearing of the neighbors of an individual, and most of the swarming systems rely on wireless swarming of states provided by external localization systems. This assumption is not realistic for real-world applications and for large swarms due to the known problems with scalability. Ideally, the individuals should rely on their sensory capabilities to estimate the relative position of their neighbors in local proximity, as it can be observed in nature. In addition, the alignment control uses the orientation of the neighbors to move in a common direction. The usage of an alignment control helps to achieve the flocking more rapidly, but it requires even more elaborate sensing mechanisms to estimate the relative orientation.

Different from formation control where more elaborate rules are designed to achieve a complex behavior \cite{Saska:2020B}, here a complex behavior is the result of local interactions between individuals through basic rules. According to \cite{Brambilla:2013}, there is still no formal or precise way to design a collective behavior. Flocking is usually based on virtual physics-based design. This approach draws inspiration from physics where each robot is considered a virtual particle that exerts virtual forces on the other robots within the group.

In our work, we apply the method published in \cite{Horyna:2022} for usage in swarms. Thus, we focus on a minimalistic distributed approach for controlling a group of \acp{uav} a continuation of work in \cite{Krizek:2022}. Thus, we performed several simulations with ten \acp{uav} and several experimental runs with nine \acp{uav}. The \acp{uav} have embedded control systems in which each UAV is capable of flight control and self-localization. The proposed approach presents a flocking control function that uses only the proximal term ($\mathcal{P}$) in order to converge and move the \acp{uav} into a unified random direction. In this scenario, we can prove that a single function can maintain cohesiveness and hold the flocking orientation unified while moving the flock of \acp{uav}. As a real-world application, we chose to experiment with a group of f450-sized \acp{uav} in a desert environment that constantly inserts disturbances into the formation through the difference in terrain level along the flight field.

Finally, we can list our contributions as:

\begin{enumerate}
    \item a \ac{3D} cohesive flocking method based only on proximal control that relies only on relative neighbor position measurement, suitable to environments with remote sensing or radio communication constraints, and irregular terrains;
    
    \item the proposed method has been also integrated with our double-layer control architecture in a manner that imitates non-holonomic behavior, which is needed for proximal control-based flocking, in real-world conditions and where each agent in the flocking uses an onboard localization method;
    
    \item to the best of our knowledge, this is the first time a nine \ac{uav} flocking experiment in a real-world environment is performed, in which each \ac{uav} has its own onboard control and localization pipeline without any centralized commands.
\end{enumerate}

\section{Related Works}
\label{sec:related_works}

Various approaches may be applied for the sake of formation maintenance, as can be seen in recent research that combines those approaches \cite{Cohen:2021}, due to their applicability regarding certain scenarios. For instance, consensus-based control and collision avoidance are usually intertwined together for the sake of reaching a collision-free consensus \cite{Nedjah:2019}. 

Among the recent works for flying robots that contributed to the state-of-the-art of swarms, it is common to rely on animal behavior. Swarms are commonly observed in nature in many birds, and insect species, which form large groups of individuals moving together toward a common target location \cite{Li:2021}. Other examples of analogous collective behavior found in animals are fish schooling and the formation of herds in ungulates \cite{Bayindir:2016, Zhao:2021}. More precise navigation, reduced energy consumption, and increased survival rate are some advantages for animals obtained through flocking. 

One of the early works about flocking was published by \cite{Reynolds:1987} in the domain of computer graphics, where the authors proposed three basic rules that are carried out by each individual in the swarm to accomplish the flocking: separation rule, cohesion rule, and alignment rule. The separation rule forces the individual to maintain a distance from their neighbors to avoid collisions. The cohesion rule allows the individual to stay close to their neighbors to keep the group together. The alignment rule aligns the individual's heading with the average heading of its neighbors.

In the past years, various mathematical models proposed to achieve the flocking behavior in multi-robot systems have the Boids \cite{Reynolds:1987} as a standard principle. As mentioned by \cite{Tan:2013}, the most common use of the rules of Reynolds in flocking is in the form of virtual forces. For example, \cite{Ferrante:2012} proposed a self-organized flocking behavior proposing a novel motion control. Different from all previous works with ground robots, where the robot moves forward with a fixed velocity and rotates based on the direction of the virtual force, their work makes the non-holonomic robot move directly forward or backward depending on the direction of the virtual force.

Initial research in the area of \ac{uav} flocking and formation flying was the study from \cite{Viragh:2014}. They proposed a decentralized control flocking algorithm for autonomous flying robots. Similar to previous works with \acp{ugv}, a repulsion and an alignment term are applied for avoiding collisions and aligning the flocking direction. For obtaining the distance and bearing of the neighbors, each UAV estimates its position using \ac{gps} and then broadcasts this information to its neighbors using local wireless communication.

\cite{Kownacki:2016} presented a decentralized control algorithm for self-organized flocking using fixed-wing \acp{uav}. Without the alignment rule, the cohesion, and separation rules proposed by \cite{Reynolds:1987} were combined with a leadership feature where one \ac{uav} is selected as the leader and is controlled by a \ac{gcs} for global flocking guidance. Analogous to the work of \cite{Viragh:2014}, the information of the neighbors was obtained by \ac{gnss} and shared using wireless communication between the individuals.

\cite{DeBenedetti:2017} developed a decentralized self-organized flocking with a focus on monitoring missions using multirotor \acp{uav}. The behavior of the flock is highly configurable by tuning a set of parameters which made the proposed method suitable for different types of monitoring missions. Each \ac{uav} follows rules inspired by Reynolds Boids models, and the position and orientation of the neighbors are also obtained using a short-range wireless communication system. Both works, \cite{Viragh:2014} and \cite{Kownacki:2016}, were tested through realistic simulations and real-world experiments while the work of \cite{DeBenedetti:2017} was tested only using realistic simulations.

\cite{Vasarhelyi:2018} proposed a decentralized \ac{uav} flocking control for confined environments. The proposed method generates a desired velocity using repulsion and velocity alignment rules. However, to maintain the individuals together, a repulsive force retains the individuals inside a bounded flight arena. In addition, a similar repulsive force describes the obstacles as a collision avoidance mechanism. The authors applied an evolutionary optimization to identify the parameters that maximize the flock speed and coherence while minimizing the collisions. Thirty identical quad-copters equipped with a Pixwawk autopilot, an onboard minicomputer Odroid C1+, an XBee module, and \ac{gnss} receivers were applied in real-world experiments in an outdoor environment. The robots used two complementary, independent, and parallel wireless modules for inter-robot communication. The XBee module broadcasts packets with a small bandwidth but with a larger range. A Wi-Fi module embedded in the onboard PC transmits packets through a local ad hoc wireless network with a large bandwidth but a shorter range. The shared data contain the geodetic position and velocity measured by the onboard \ac{gnss} receivers, and the relative information comes from the differences in \ac{gnss}-based absolute measurements.

\cite{Silic:2019} introduced an atmospheric platform system for plume monitoring using fixed-wing \acp{uav}. Since the arrangement of the sensors in the environment is crucial to obtaining rich information, the main objective was to organize the UAVs equipped with onboard sensors in a cohesive formation inside the plume boundary. To explore an unknown environment, the \ac{uav} randomly moves to a fixed position and gathers data for a while. The \ac{gcs} receives from the \acp{uav} the environmental data through wireless communication and then processes it to estimate the plume location. Also, the \ac{gcs} organizes the formation by receiving their \ac{gps} coordinates and sending new positions to investigate. Once the position of the plume is well-known, they keep loitering in circles inside the plume boundary. The system was evaluated with three delta wing \acp{uav} with a \ac{gps} receiver, a radio transceiver, a barometer, an atmospheric sensor, and a 9-axis inertial measurement unit. The \ac{uav} communicates only with the \ac{gcs} using XBee radios.

Another common problem in robotic swarms is the path planning of \ac{uav} in a \ac{3D} environment. This problem tackles an important part of the entire autonomous control system of the \ac{uav}. In the constrained mission environment, planning optimal paths for multiple \acp{uav} is a challenging problem. To solve this problem, \cite{He:2021} proposed a \ac{TSS} model to simplify the handling of coordination cost of \acp{uav}. They also proposed a novel hybrid algorithm by combining an \ac{ipso} and a \ac{msos}, called HIPSO-MSOS.

Meta-heuristic approaches are very common in swarm research. An example can be found in the work of \cite{Dentler:2019}. In \cite{Dentler:2019}, a novel mobility model combining an \ac{CACOC} was proposed in order to enhance the area coverage of such \ac{uav} swarms. The authors extended the \ac{CACOC} approach by a collision avoidance mechanism and tested its efficiency in terms of area coverage by the UAV swarm. Another approach can be found in the work of \cite{Phung:2021}, where they present an algorithm named \ac{SPSO} to deal with the problem of path planning \acp{uav} in complicated environments subjected to multiple threats.

In \reftab{tab:comparison_flying}, we compare the mentioned UAV flocking algorithms and our proposed work. Flocking is often accomplished with decentralized planning using GNSS information. The work of \cite{Viragh:2014}, \cite{Kownacki:2016}, and \cite{Vasarhelyi:2018} were tested through realistic simulations and real-world experiments while the work of \cite{DeBenedetti:2017} was tested only using realistic simulations. With experiments using thirty physical UAVs, the work of \cite{Vasarhelyi:2018} can be considered one of the most impressive achievements so far. However, achieving coordinated swarm behaviors without external sensing and computation is a challenging task, as has been very well explained in the recent survey of \cite{Coppola:2020}.

\begin{table*}[ht]
    
            \centering
            \caption{Comparison between the mentioned UAV flocking algorithms and our work.}
            \label{tab:comparison_flying}
            \begin{tabular}{@{}lccc@{}}
                \toprule
                Work & Decentralized control & Real-world experiment & Local sensing \\
                \midrule
                \cite{Viragh:2014} & X & X &  \\
                \cite{Kownacki:2016} & X & X & \\
                \cite{DeBenedetti:2017} & X & & \\
                \cite{Vasarhelyi:2018} & X & X & \\
                \cite{Silic:2019} & & X &  \\
                Our work & X & X & X \\
                \bottomrule
            \end{tabular}
\end{table*}

The current state-of-the-art discussed in this section presents relevant contributions and has provided valuable foundations. However, to the best of our knowledge, our work is the first method based only on proximal control for ordered and cohesive \ac{2D}/\ac{3D} flocking in real \ac{uav} environments applied in either \ac{gnss} or \ac{gnss}-denied environments. 

\section{UAV Model application}
\label{sec:uav_model}


Multirotor \acp{uav} may be considered underactuated vehicles with holonomic mobility. When designing a reliable decentralized flocking mechanism, we have to assume that the UAVs are subjected to localization problems with an error of position estimation by GPS in the order of meters. Communication of the relative position between UAVs within the flock also suffers drawbacks due to communication delays, intermittent communication interruptions, and mainly scalability to enable the operation of large swarms, which is the main aim of the swarm community. Nevertheless, the applied control \ac{mrs} (MRS) group \ac{uav} system \cite{Baca:2021} allows for both holonomic and non-holonomic movements. Thus, let us use the dynamical model of a multirotor aerial vehicle presented in the work of \cite{Lee:2010}.

\begin{figure}[t]
    \centering
    \includegraphics[width=0.45\textwidth]{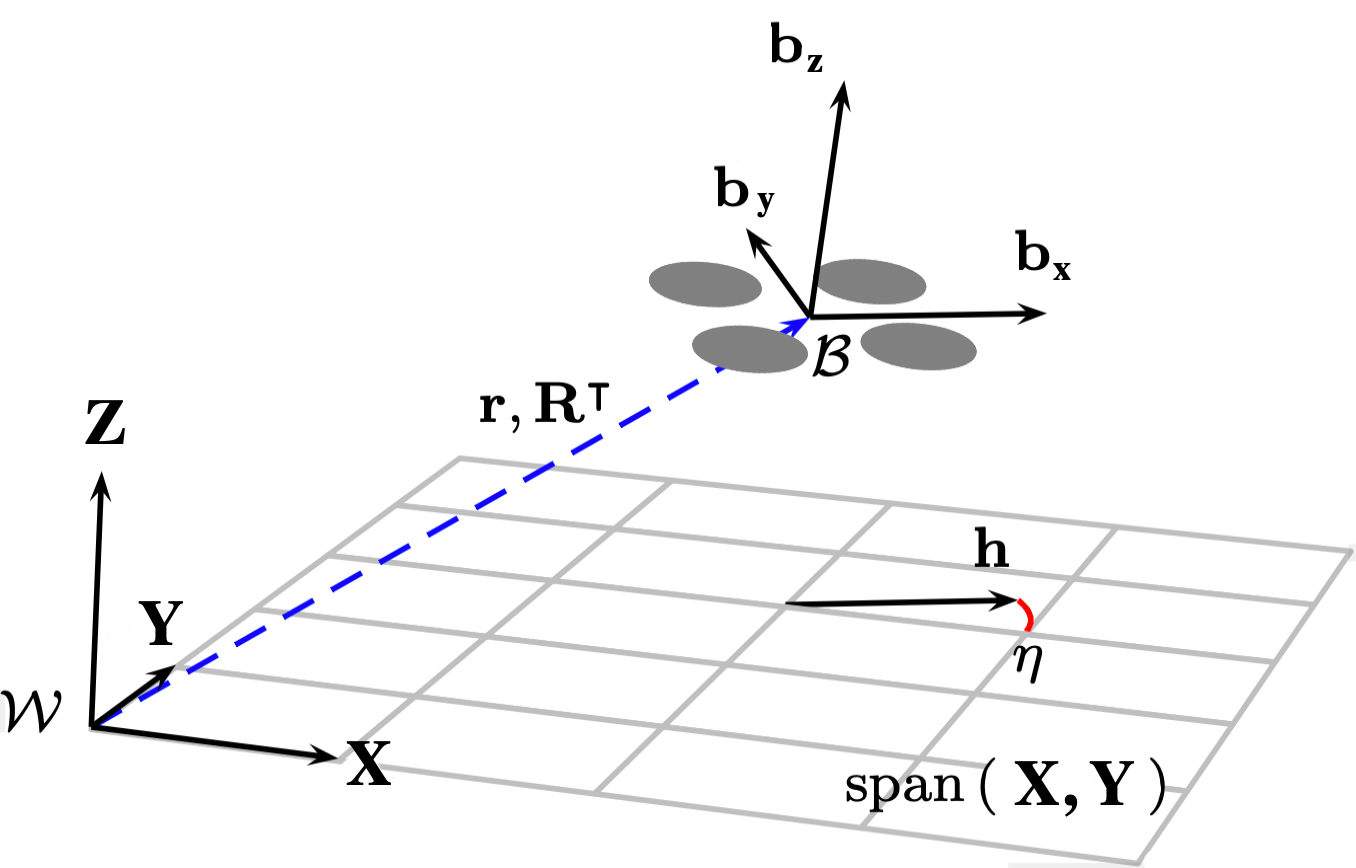}
    \caption{UAV Model schematics.}
    \label{fig:coordinate_frames}
\end{figure}

To better analyze this model, let us consider \reffig{fig:coordinate_frames}. Here, we depict the world frame $\mathcal{W}$ = $\{X, Y, Z\}$ in which the 3D position and the orientation of the UAV body are expressed. The body frame $\mathcal{B}$ = $\{\mathbf{b}_x$, $\mathbf{b}_y$, $\mathbf{b}_x\}$ relates to $\mathcal{W}$ by the translation $\mathbf{r} = \left[x, y, z\right]^{\intercal}$ and by rotation $\mathbf{R}^{\intercal}$. The UAV heading vector $\mathbf{h}$, which is a projection of $\mathbf{b}_x$ to the plane $span\left(X, Y\right)$, forms the heading angle $\eta$. Thus, the UAV nonlinear model can be expressed as having a translation part:

\begin{equation}
    m\ddot{\mathbf{r}} = f_T\mathbf{R}Z - m g Z,
    \label{eq:model_UAV_translation}
\end{equation}
\noindent where $m$ is the \ac{uav} mass, $g$ is the gravity acceleration, $f_T$ is the thrust force created by the propellers in the direction of $\mathbf{b}_z$, $\mathbf{r}=\left[x, y, z\right]^\intercal$ is the position of the center of the mass of a UAV in the world frame, $\ddot{\mathbf{r}} \in \real^3$ is the acceleration of the center of the mass of a UAV in the world frame, ${\mathbf{R}} \in \mathrm{SO}(3) \subseteq \real^{3\times3}$ is the rotation matrix from the body frame of a \ac{uav} to the world frame, and a rotational part:

\begin{equation}
    \dot{\mathbf{R}} =\Omega\mathbf{R} ,
    \label{eq:model_UAV_rotation}
\end{equation}
\noindent where $\Omega$ is the tensor of angular velocity, under the condition $\Omega\,\mathbf{\dot{z}}= \dot{\theta} \times\mathbf{\dot{z}}, \forall \mathbf{\dot{z}} \in Z$. The \ac{uav} is affected by downward gravitational acceleration with $g \in \mathbb{R}$.

Furthermore, we separately consider and estimate the azimuth of the $\mathbf{b}_x$ axis in the world as the \ac{uav} \emph{heading}. Under the condition of $\mid Z^\intercal\mathbf{b}_x \mid > 0$, we define the heading as:

\begin{equation}
    \eta = \mathrm{atan2}\left(\mathbf{b}_x^\intercal Y, \mathbf{b}_x^\intercal X\right) = \mathrm{atan2}\left(\mathbf{h}_{(y)}, \mathbf{h}_{(x)}\right).
\end{equation}

The heading is a more intuitive alternative to the widely-used \emph{yaw} angle as one of the 4 controllable \acp{dof}. Furthermore, it is possible to use the $yaw$, but for that, we would need to assume that the tilt of the UAV ($\mathrm{cos}^{\minus 1}\ (\mathbf{b}_z^{\intercal}Z)$) is low, near horizontal. Thus, we define the heading vector by the $\mathbf{b}_x$ axis as:

\begin{equation}
    \mathbf{h} = \left[\mathbf{R}_{(x,x)}, \mathbf{R}_{(y,x)}, 0\right]^\intercal = \left[\mathbf{b}_x^\intercal X, \mathbf{b}_x^\intercal Y, 0\right]^\intercal.
\end{equation}

And its normalized form:
\begin{equation}
    \mathbf{\hat{h}} = \frac{\mathbf{h}}{\|\mathbf{h}\|} = \left[\mathrm{cos}\,\eta, \mathrm{sin}\,\eta, 0\right]^\intercal.
\end{equation}

Therefore, we can note that all these factors directly influence the estimate of range and bearing between neighboring robots, the bearing estimation being the most problematic in teams of UAVs. Thus, when implementing this algorithm into multirotor \acp{uav}, one must first decrease the robot dynamics to stabilize the flocking (i.e., similar to mimicking the friction effect). \acp{uav} are highly unstable and tend to drift due to errors in state estimation. Furthermore, in our approach, the robots behaved in a non-holonomic manner. This helps stabilize the UAV, minimizing rapid variations in position by moving sideways. Although the rotation dynamics are considered in the inner control loop \cite{Baca:2021}, our approach avoids sending such reference in order to maintain the heading to a desired fixed value.

\section{UAV Flocking Approach}
\label{sec:uav_flocking}

Multi-robot group control methods based on potential functions are generally composed of three different term functions: a proximal term, an alignment term, and an optional goal direction term, which is needed when the swarm is required to steer toward a specific target. In our work, we propose a flocking control function that uses only the proximal term ($\mathcal{P}$) to converge and move the UAVs in a unified direction.

\begin{definition}
    In a flocking of $n$ robots, a $i^{th}$ robot, where $i=1..n$, is called the focal robot $F_r$. 
\end{definition}

Thus, the proximal term function $\mathcal{P}$ of the focal robot $F_r$ results in a vector that enables the focal robot to maintain the desired distance from other neighbor robots while keeping the group cohesiveness. The proximal term encapsulates both the attraction and the repulsion behaviors, commonly seen in the potential fields and other swarming approaches \cite{Brambilla:2013}. The proximal term function is widely used to achieve cohesive flocking. By using this function, a robot maintains the desired distance from other neighboring robots while keeping a cohesive formation. By applying the proximal term in the \ac{2D} space, the range, and bearing is the only information required to compute this term. Thus, to calculate $\mathcal{P}$ we first must estimate the relative pose of each neighbor, which in turn is required for obtaining the relative range $d_n$ and bearing $\phi_n$, and it is expressed in the body frame of reference of the focal robot.

\begin{assumption}
    In a flocking of $n$ robots, the $n^{th}$ neighbor is a robot within the sensor range $d_{Lim}$ delimited by the interaction range gain $\lambda$, where $d_{Lim}=d_{des}\lambda$, with $d_{des}$ being the desired distance between robots in the flocking. Thus, the range $d_n$ of the $n^{th}$ neighbor, which the focal robot $F_r$ takes into account, is such that $d_n \leq d_{Lim}$.
    \label{asm:flocking}
\end{assumption}

\subsection{Direct Information Exchange for Intra-swarm Relative Sensing}
\label{subsec:direct_info}

In outdoor environments, relative position is obtained via a combination of \ac{gnss} and intra-swarm communication in most of the existing swarming approaches. Global position information obtained via \ac{gnss} is communicated between \acp{mav} and then used to extract relative position information \cite{Coppola:2020}. For each global position information received through communication, within the flocking, a focal robot calculates relative position as:

\begin{equation}
    \begin{aligned}
        x_r = x_n - x_f, \\
        y_r = y_n - y_f, \\
    \end{aligned}
    \label{eq:pos_relative_2d}
\end{equation}
\noindent where $x_n$ is the coordinate of the $n^{th}$ neighbor in the $X$-Axis, $x_f$ is the coordinate of the focal robot ($F_r$) in the $X$-Axis, $x_r$ is the relative x-coordinate of the neighbor robot on the body frame of $F_r$, $y_n$ is the coordinate of the $n^{th}$ neighbor in the $Y$-Axis, $y_f$ is the coordinate of the focal robot ($F_r$) in the $Y$-Axis, and $y_r$ is the relative y-coordinate of the neighbor robot on the body frame of $F_r$. 

For the \ac{3D} flocking of \acp{uav} using \ac{gps} and broadcasting their positions to the neighbors, the coordinate of the neighbor in the $Z$-Axis is calculated on the body frame of $F_r$. When using the approach in the \ac{2D} space, this information is used only to maintain all \acp{uav} at the same height. A \ac{3D} case is usually either stimulated by initializing the \acp{uav} in different altitudes or when an external ground irregularity forces the \ac{uav} to be in a \ac{3D} shape. When estimating the height of $F_r$ using the Garmin Lidar, the effect of this behavior is that when one UAV goes up to avoid an irregular terrain, the others follow it also to avoid the bump in the terrain even without seeing it. Thus, the coordinate of the neighbor in the $Z$-Axis on the body frame of $F_r$ is calculated as:

\begin{equation}
    z_r = z_n - z_f, 
    \label{eq:pos_relative_height}
\end{equation}
\noindent where $z_n$ is the coordinate of the $n^{th}$ neighbor in the $Z$-Axis, $z_f$ is the coordinate of the focal robot in the $Z$-Axis, and $z_r$ is the coordinate of the neighbor robot in the $Z$-Axis on the body frame of $F_r$. 

\subsection{UVDAR-Based Intra-swarm Relative Sensing}

On many occasions, we are incapable of retrieving the information on global position when the \acp{uav} are in \ac{gnss}-denied environments or implicit communication is not allowed. Therefore, we have to use the local origin coordinate frame. The local origin frame is the coordinate frame with center and orientation coincident with the starting point and orientation of the \ac{uav}. The usage of the local origin frame will only provide the correct estimation if all the UAVs start at the same height. This is explained by the fact that the height estimation is performed by the Garmin Lidar in each UAV, and thus, this measurement does not account for differences in the level of the terrain. Using the onboard \ac{uvdar} system for relative localization of neighbors, the transformations discussed previously are not necessary since the \ac{uvdar} sensor provides the pose of the neighbor in the body frame of the robot, requiring only the transformation to the body frame of the focal robot coordinate frame.

\begin{figure}[t]
  \centering
  \includegraphics[width=0.475\textwidth]{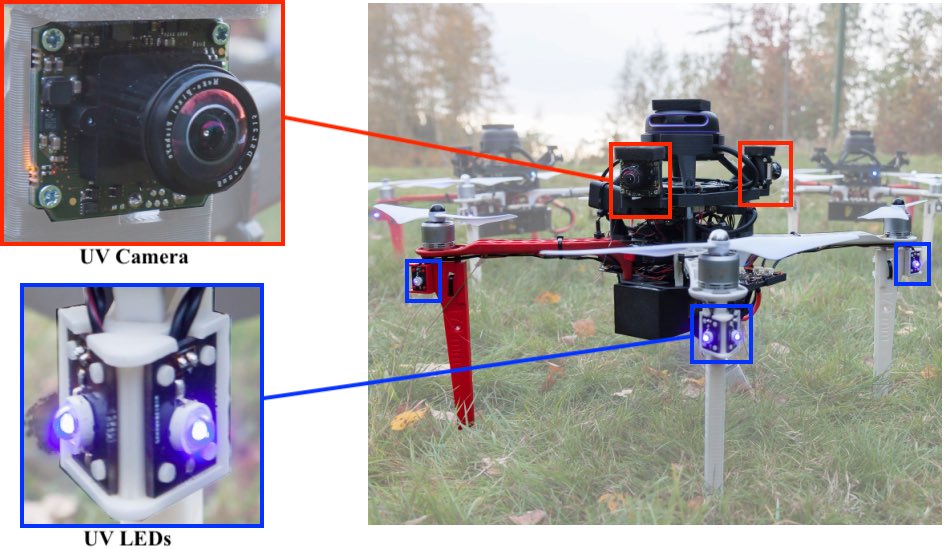}
  \caption{\ac{uvdar} \acp{led} and cameras.}
  \label{fig:uvdar_cameras}
\end{figure}

The \ac{uvdar} system is based on the application of markers composed of \ac{UV} \acp{led}, and cameras with fish-eye lenses and specialized band-pass filters carried by all \acp{uav} of the swarm \cite{Walter:2019} (see \reffig{fig:uvdar_cameras}). To be recognized by the other teammates, each \ac{uav} uses four pairs of blinking \ac{UV} markers attached to its arms. For recognizing the others, a \ac{uav} uses two modified cameras with fish-eye lenses allowing almost $360^{\circ}$ horizontal \ac{fov}.  Using a properly calibrated camera, the \ac{uvdar} system retrieves from images the distance of a neighbor UAV using a triangulation between the fish-eye camera and at least two \acp{led}, and then calculates its relative position.

The setup of the \ac{uvdar} system with two cameras allows a \ac{2D} flocking with \acp{uav} staying at the same height that keeps the \ac{uvdar} \acp{led} of the neighbor robots in the range of view of the \ac{uvdar} cameras of the focal robot making it easier to estimate the position of the neighbors. An omnidirectional \ac{uvdar} system requires three to four cameras to achieve the same performance for \ac{3D} flocking.

\subsection{Magnitude-dependent Flocking Motion Control}

Independent of the sensor used (\ac{gnss} or \ac{uvdar}), we can calculate the range of the $n^{th}$ neighbor $d_n$ as:
\begin{equation}
    d_n = \sqrt{x^2_r + y^2_r + z^2_r}.
    \label{eq:3d_range}
\end{equation}

\begin{figure}[t]
    \centering
    \includegraphics[width=0.5\textwidth]{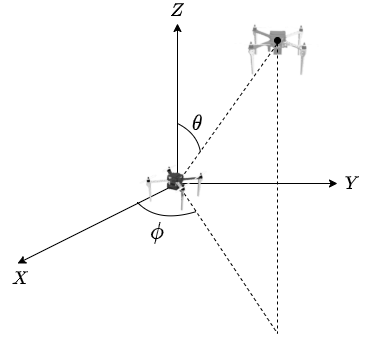}
    \caption{Representation of the bearing $\phi$ and the inclination $\theta$ in the body frame of the focal robot.}
    \label{fig:3d_representation}
\end{figure}

To calculate the bearing of the $n^{th}$ neighbor in \ac{3D} we have to consider that there are two different angles to take into account (see \reffig{fig:3d_representation}). The first angle is the normal bearing $\phi_n$. The bearing $\phi_n$ of the $n^{th}$ neighbor can be obtained in the XY plane as:

\begin{equation}
    \phi_n = atan2(y_r, x_r).
    \label{eq:bearing}
\end{equation}

For a general \ac{3D} flocking, the angle between the positive $Z$-Axis and the line segment between the origin of the focal coordinate frame to its $n^{th}$ neighbor position $\theta_n$ needs also to be obtained. The addition of $\theta_n$ to the proximal term function makes it possible for each \ac{uav} to ascent or descent to keep a cohesive swarm while flocking on uneven terrain. The angle $\theta_n$ is calculated as:

\begin{equation}
    \theta_n = \begin{cases}
        0 & \text{if } z_r = 0, \\
        atan2(\sqrt{x^2_r + y^2_r}, z_r) & \text{otherwise}.
    \end{cases}
    \label{eq:inclination}
\end{equation}

After estimating the range and bearing of each neighbor, the proximal term is calculated as:
\begin{equation}
    \mathcal{P}=\sum^{m_p}_{n=1} p_n(d_n)e^{i\phi_n + j\theta_n},
    \label{eq:p_term_3d}
\end{equation}
\noindent where $m_p$ is the number of neighboring \acp{uav} perceived by the focal robot within the maximum interaction distance $d_{Lim}$, and where the term $p_n(d_n)$ is the magnitude of the proximal vector.

The term $p_n(d_n)$ can be calculated as a Lennard-Jones potential function such that \cite{Ferrante:2012}:

\begin{equation}
    p_n(d_n)=-\frac{\partial P(d_n)}{\partial d_n} = -\frac{4\alpha \epsilon}{d_n}\left[2\left(\frac{\sigma}{d_n}\right)^{2\alpha} - \left(\frac{\sigma}{d_n}\right)^{\alpha}\right],
    \label{eq:p_derivative}
\end{equation}
\medskip\noindent where $P(d_n)$ is a virtual potential function, $\epsilon$ is the strength of the potential function which determines its depth, $\alpha$ is the steepness of the potential function, and $\sigma$ is the amount of noise calculated by equation (\ref{eq:pf_minimum}). Thus, the amount of noise $\sigma$ can be expressed as:
\begin{equation}
  \sigma=\frac{d_n}{2^{\frac{1}{\alpha}}}.
  \label{eq:pf_minimum}
\end{equation}

Note that the minimum value of $\epsilon$ should be when $d_n=d_{des}$. To achieve the advantages of magnitude-dependent flocking motion control, all UAVs need to satisfy non-holonomic constraints similar to the constraints of differential drive robots. But in order to pass the information needed by the UAV control system, we need first to decompose the flocking control vector $\mathcal{P}$ into a motion control vector. We can first decompose the value of $\mathcal{P}$ into three components: $p_x$, $p_y$, and $p_z$. We call $p_x$, $p_y$, and $p_z$ the projection of the flocking control vector $\mathcal{P}$ on the XYZ-plane of the body reference frame of the focal robot, obtained as:

\begin{equation}
    \begin{aligned}
        &p_x = \sum^{m_p}_{n=1} p_n(d_n)\sin{\theta_n}\cos{\phi_n}, \\
        &p_y = \sum^{m_p}_{n=1} p_n(d_n)\sin{\theta_n}\sin{\phi_n}, \\
        &p_z = 
        \begin{cases}
          0 & \text{if } z_r = 0, \\
          \sum^{m_p}_{n=1} p_n(d_n)\cos{\theta_n} & \text{otherwise}. 
        \end{cases}
    \end{aligned}
    \label{eq:3d_projection}
\end{equation}

The next step is to convert the projections $p_x$, $p_y$, and $p_z$ into a desired movement. We start by multiplying each projection by a gain as:

\begin{equation}
    \begin{aligned}
        &u = \kappa_1 p_x + B_s, \\
        &v = \kappa_2 p_y, \\
        &\omega = \kappa_3 p_z. 
    \end{aligned}
    \label{eq:movement}
\end{equation}
\noindent where $k_i$, with $i \in \{1,\ldots, 3\}$, are the proportional gains to transform from the respective projections into the additive speed component, and $B_s$ is the maximum forward biasing speed in the X-Axis.


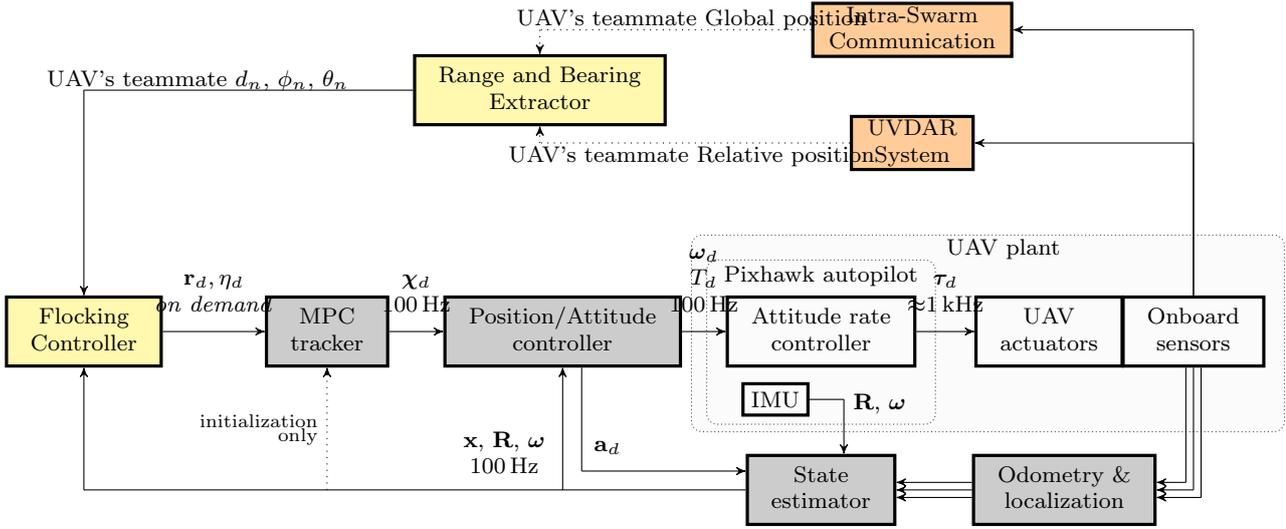
\begin{figure*}[t]    
    \input{fig_system_archicteture}
    \caption{
      A diagram of the system architecture: the \emph{Flocking Controller} and the \emph{Range and Bearing Extractor} in yellow are what we propose in this work. Both blocks are built upon the \ac{mrs} system software (blocks in gray) \cite{Baca:2021} and supply the desired reference (position $\mathbf{r}_d$ and heading $\eta_d$) to the \ac{mrs} system. In this diagram, the white blocks outline the physical design of the \ac{uav} while the orange blocks outline the relative localization approaches (either using GPS and communication through Wi-Fi or using \ac{uvdar} \cite{Walter:2019}). Within the \ac{mrs} system, we have a first layer with a \ac{mpc} tracker that processes the desired reference and gives a full-state reference to the position/attitude controller. \emph{MPC tracker} creates a smooth and feasible reference $\bm{\chi}$ for the reference feedback controller. The feedback \emph{Position/Attitude controller} produces the desired thrust and angular velocities ($T_d$, $\bm{\omega}_d$) for the Pixhawk embedded flight controller (Attitude rate controller). The \emph{State estimator} fuses data from \emph{Onboard sensors} and \emph{Odometry \& localization} methods to create an estimate of the UAV translation and rotation ($\mathbf{x}$, $\mathbf{R}$).
    }
    \label{fig:system_architecture}
\end{figure*}

Here, $u$ is assumed to be directly proportional to the $x$ projection of the vector (i.e., $p_x$), which results in the forward movement of the robot in the body reference frame, $v$ will be directly proportional to the $y$ projection of the vector (i.e., $p_y$) and $\omega$ will be directly proportional to the $z$ projection of the vector (i.e., $p_z$). 

Finally, to send to the dual-layer control system of the \ac{uav} the desired position $\mathbf{r}_d = \left[x_d, y_d, z_d\right]^\intercal$ and heading $\eta_d$ in the global frame as input values, we must update its current position and heading. Thus, we first use $u$ as:

\begin{equation}
    \begin{aligned}
        &x_d(k+1) = x_d(k) + \cos{\eta_d(k)} u, \\
        &y_d(k+1) = y_d(k) + \sin{\eta_d(k)} u .
    \end{aligned}
    \label{eq:3d_pos_update}
\end{equation}

Note in equation (\ref{eq:3d_pos_update}) that the larger the value of $p_x$, the faster the robot moves forward, the larger the value of $p_y$, and the faster the robot turns, and the larger the value of $p_z$, the faster the robot moves up or down. Although this approach allows the robot to move backward, it tends to move forward due to the existence of the forward biasing movement of $B_s$.

In contrast, the $Z$-Axis is a special case. We must first calculate the needed height gain $h_{push}$, by taking into account a threshold of allowed height $h_{limit}$ for the case the UAV goes below the minimum height. Thus,

\begin{equation}
    \begin{aligned}
        &h_{push} = \max(0, h_{limit}-z_f).
    \end{aligned}
    \label{eq:Hpush}
\end{equation}

Thus, the initial desired height $h_d$ is then calculated using $\omega$ as follows:

\begin{footnotesize}
\begin{equation}
    \begin{aligned}
        &h_d(k) = 
        \begin{cases}
          z_d(k) + h_{push} & \text{ if } (z_f + h_{push} + v) \leq h_{limit}, \\
          z_d(k) + h_{push} + \omega  & \text{ otherwise}.
        \end{cases}
    \end{aligned}
    \label{eq:desired_height}
\end{equation}
\end{footnotesize}

Finally, we must take into account if we require achieving a \ac{2D} or \ac{3D} formation, respectively. Thus, the final reference height $z_d$ is as follows:

\begin{equation}
    \begin{aligned}
        &z_d(k+1) = 
        \begin{cases}
            h_d(k) + z_n & \text{ if } z_r = 0, \\
            h_d(k) & \text{ otherwise}. 
        \end{cases}
    \end{aligned}
    \label{eq:desired_height2}
\end{equation}

The last part of the flocking algorithm is regarding the heading of the UAV. By using the $v$ value, we can calculate the UAV desired reading ($\eta_{des}$) as follows:

\begin{equation}
    \begin{aligned}
        &\eta_{des} = \eta_d(k) + v,
    \end{aligned}
    \label{eq:3d_head_update}
\end{equation}
\noindent where $\eta(k)$ is the current heading of the focal robot at the instant $k$.

However, we must also take into account the \ac{uav} heading rapid variation before sending the desired references to the dual-layer control system from the \ac{mrs} system \cite{Baca:2021}. Even when stabilizing the \ac{uav}, the presence of rapid variation on the bearing caused by high forces due to big differences in angles still presents a major problem. In order to minimize this problem, we re-calculate the observed bearing with a \ac{SLERP} technique. \ac{SLERP} is a popular technique for interpolating between two 3D rotations while producing smooth paths. Thus, the recalculation of the bearing transforms the robot's physical heading angle:

\begin{small}
    \begin{equation}
        q_S(k+1) = \left[\frac{\sin((1-\gamma)\psi)}{\sin{\psi}}\right]q_{\eta}(k)+\left[\frac{\sin(\gamma \psi)}{\sin{\psi}}\right]q_S(k),
        \label{eq:slerp_1}
    \end{equation}
\end{small}
\noindent where $q_S (k)$ is the quaternion of the smoothed reference heading angle ($\eta_d$) in the instant $k+1$, $q_{\eta}$ is the quaternion of the physical desired heading angle ($\eta_{des}$), $\psi$ is the angle between the two quaternions, and $\gamma$ is an interpolation coefficient. For $\psi \approx 0$, the \ac{SLERP} equation is:

\begin{equation}
  q_S(k+1) = (1-\gamma)q_{\eta}(k)+\gamma q_S(k),
 \label{eq:slerp_2}
\end{equation}
\noindent and through quaternion $q_S(k+1)$ we find $\eta_d(k+1)$, which is the smoothed final reference heading angle. 

Finally, note that to be able to achieve self-organized ability towards the intended collective motion, we applied motion constraints protecting against unwanted side-motion of \acp{uav}. In comparison with \acp{ugv}, \acp{uav} are highly unstable and tend to drift due to errors in state estimation. The control architecture composed of a \emph{\ac{mpc} tracker} coupled with the \emph{SE(3) geometric state feedback} position/attitude controller based on our previous work \cite{Baca:2021} was therefore used.

\subsection{Control Architecture}

The control architecture used in this work consists of several interconnected subsystems, as depicted in \reffig{fig:system_architecture}. The \emph{Flocking} block embeds the above-discussed approach, the main contribution of this paper, and supplies the time-parametrized sequence of the desired position and heading. This information is then sent to a dual-layer control system formed by a tracker and a position/attitude controller. Thus, at first, the proposed \emph{Flocking} approach sends the reference position and heading to the \emph{\ac{mpc} tracker} \cite{Baca:2018}, which in turn provides a full-state of the UAV to the position/attitude SE(3) controller \cite{Lee:2010}. At the \emph{\ac{mpc} tracker} block, the desired position, and heading are converted into a feasible, smooth, and evenly-sampled full-state control reference. This control reference contains the desired position, its derivatives up to the jerk, the heading, and the heading rate, supplied at \SI{100}{\hertz}. Then, the full-state control reference is used by a \emph{SE(3) geometric state feedback controller} to provide feedback control of the translational dynamics and the orientation of the UAV. This block creates an attitude rate $\bm{\omega}_d$ and a thrust command $T_d$, which are sent to an embedded flight controller\footnote{Pixhawk is used for experimental verification.}. The flight controller encapsulates the underlying physical UAV system with motors and motor ESC and creates 4 new controllable \acp{dof}: the desired angular speed around ($\mathbf{\hat{b}}_1, \mathbf{\hat{b}}_2, \mathbf{\hat{b}}_3$) and the desired thrust $\left<0, 1\right>$ of all propellers. Finally, onboard sensor data (e.g., position measurements from Barometer/Garmin Lidar, velocity measurements from visual odometry, etc.) are processed by the \emph{Odometry \& Localization} and the \emph{State estimator} blocks and supply the needed information to all these previous blocks.

\begin{table}[t]
    \centering
        \caption{Parameter values} \label{tab:parameters}
        \begin{tabular}{@{}ccc@{}}
                \toprule
                Parameter & Description & Value \\
                \midrule
                $B_S$ & Maximum forward speed & 0.3 m/s \\
                $\kappa_1$ & Linear gain & 0.5 \\
                $\kappa_2$ & Angular gain & 0.2 \\
                $\kappa_3$ & Height gain & 0.1 \\
                $\alpha$  & Steepness of potential function & 2 \\
                $\epsilon$ & Strength of potential function & 6 \\ 
                $d_{des}$  & Desired inter-robot distance & 6 m \\
                $D_p$  & Maximum interaction range & 7.2 m \\
                $\lambda$  & Interaction range gain & 1.8 \\
                $\gamma$  & Interpolation coefficient & 0.95 \\
                $h_{limit}$  & Minimum desired height & 2.0 m \\
                $T$ & Duration of experiments & 160 s \\
                \bottomrule
            \end{tabular}
\end{table}

\section{Results}
\label{sec:results}

Swarming algorithms are intended for large groups of robots ($\ge 100$). When using ground robots, gathering this high number of robots is somehow feasible. However, in our experiments, we aim to analyze the behavior of autonomous middle-size UAVs, where real disturbances from outdoor environments can disturb group cohesiveness. In this case, the use of numerous \acp{uav} is not a simple task. Thus, we performed real robot experiments with ten robots and therefore succeeded in proving the efficiency of our proposed approach. 

The aim of the experiments presented here is to verify local interactions between UAVs without the necessity of sharing position information among the robots in the fleet. We aim to show whether the flocking mechanism for UAVs can achieve similar properties as in the work on \acp{ugv} \cite{Ferrante:2012}, where scalability to larger swarms was proven. The experiments used the parameters observed in \reftab{tab:parameters}.

\subsection{Metrics}

To analyze the effectiveness of our approach, we use two metrics. The first, the order metric $\psi$,  measures the degree of alignment of the orientations within the swarm. A swarm having a common heading will result in a value of $\psi \approx 1$. A swarm where the \acp{uav} are pointing in different directions will result in a value of $\psi \approx 0$. Using the vectorial sum of the headings of all $N$ robots, the order metric $\psi$ is defined as:
\begin{equation}
  \psi = \frac{1}{N} \Vert\sum^{N}_{i=1} e^{j\phi_i}\Vert.
  \label{eq:order_metric}
\end{equation}

We also analyze the steady-state value that is reached for a given metric (in our case the order metric $\psi$). The steady-state metric $\Bar{\mu}$ is the asymptotic value reached by the order metric during the experiment. After the system reaches a steady state, we calculate the steady-state value as the average value of the metric in the last 100 seconds. More formally:
\begin{equation}
  \Bar{\mu} = \frac{\sum^{T}_{(t=T-100)} \psi_t}{100},
  \label{eq:steady_metric}
\end{equation}

\noindent where $T$ is the time duration of the experiment.

\subsection{Real Robot Experiments}

For the real robot experiments, we used nine identical DJI f450-based quadcopters. Each quadcopter uses a Pixhawk 4 flight controller, an onboard Intel NUC7 PC with Linux Ubuntu, and a Tersus GPS receiver with a \ac{gnss} system. The communication between the UAVs is supported by the Wi-Fi module embedded in the PC. For the experiments with the \ac{uvdar} system, each UAV is also equipped with the UV-cameras and \ac{UV}-\acp{led} required to recognize and be perceptible by, the other teammates.

The experiments were performed in the desert of Abu Dhabi with dunes up to two and a half meters high. The environment had moderate high winds with speeds up to 12km/h. We performed 10 runs with each sensory system. One of each run was presented here as a result. Because the expected behavior of the swarm is a constant forward movement in an arbitrary direction, the experiments with real robots were performed in a delimited area with the \acp{uav} almost facing the same direction (with small variations in angles). As a safety measure, the whole experiment finishes if any individual reaches any of the area borders. We kept the same number of robots for each experiment: nine robots when using the direct information exchange-based approach and nine robots when using the \ac{uvdar} system.

\begin{figure}[!ht]
    \centering
    \includegraphics[width=0.45\textwidth]{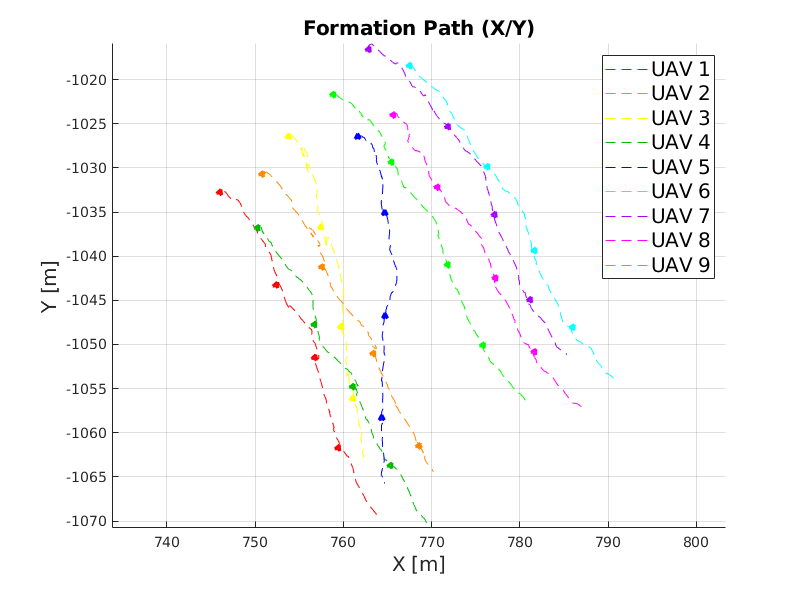}
    \caption{XY plot showing UAV headings during the experiment with real robots using the direct information exchange-based approach.}
    \label{fig:real_3d_gps_flocking}
\end{figure}

\begin{figure}[!ht]
    \centering
    \includegraphics[width=0.45\textwidth]{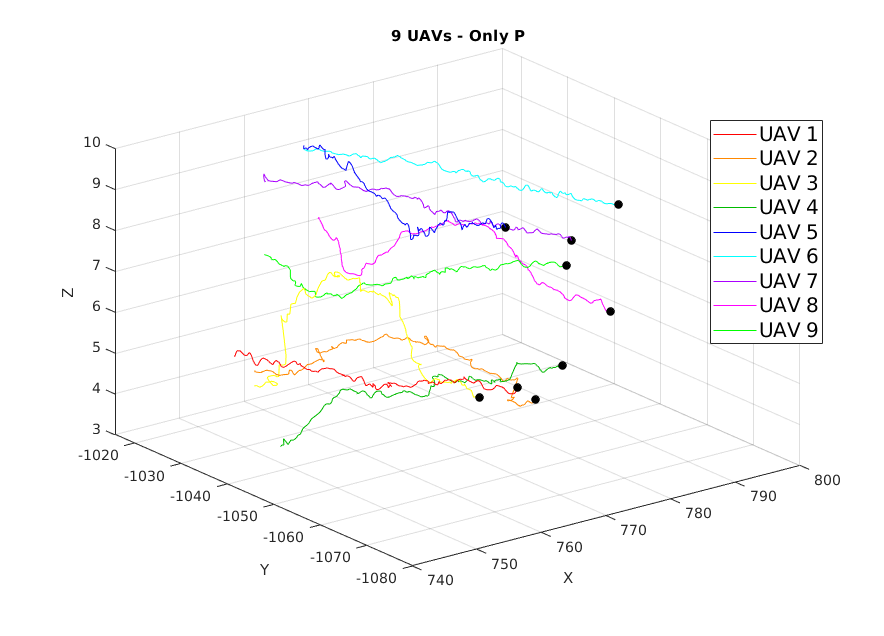}
    \caption{The 3D plot of the UAV flocking.}
    \label{fig:real_gps_flocking}
\end{figure}

\subsubsection{Results with Direct Information Exchange-based approach}

After the take-off and hovering for a 10s period, the experiment with nine \acp{uav} starts. With only the proximal approach proposed here, the UAVs rapidly converged in range and bearing. We selected this run of experimental evaluation to show the behavior of the flocking with several autonomous robots in a distributed flocking system. Usually, there is an inherent inaccuracy in using a \ac{gnss} sensor of approximately 1.5 meters. This can lead to a spread formation. After a while, as Lennard-Jones' potential function does not present strong forces in aerial systems, the formation with numerous \acp{uav} can break into small groups (see \reffig{fig:real_3d_gps_flocking}). However, even with this error in the position estimate, the flocking algorithm was successful in converging the robots into a cohesive \ac{3D} group (see \reffig{fig:real_gps_flocking}). 

\begin{figure}[t]
  \centering
  \includegraphics[width=0.5\textwidth]{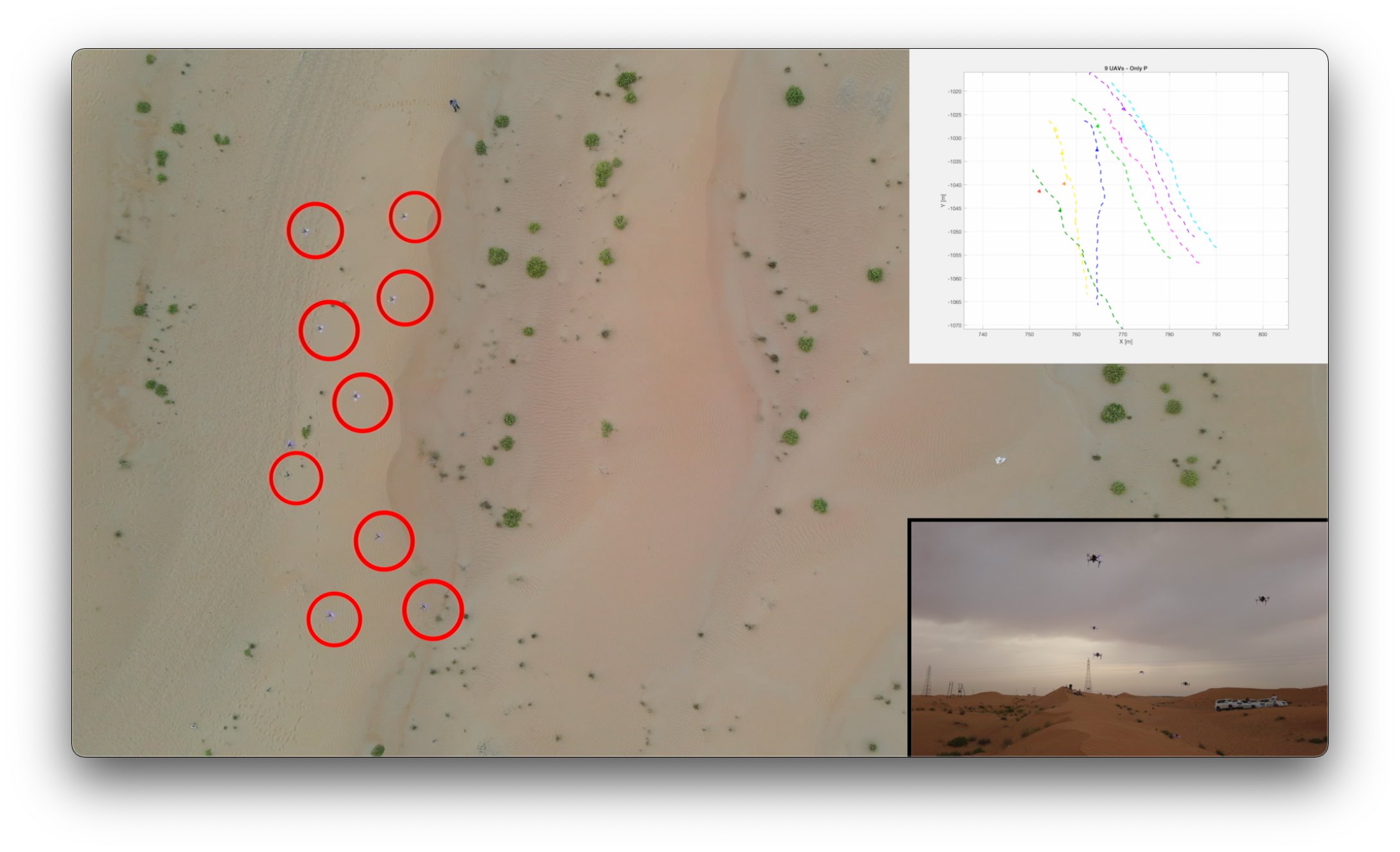}
  \caption{Real robot UAV 3D flocking.}
  \label{fig:real_gps_heading}
\end{figure}

\begin{figure}[!ht]
  \centering
  \includegraphics[width=0.5\textwidth]{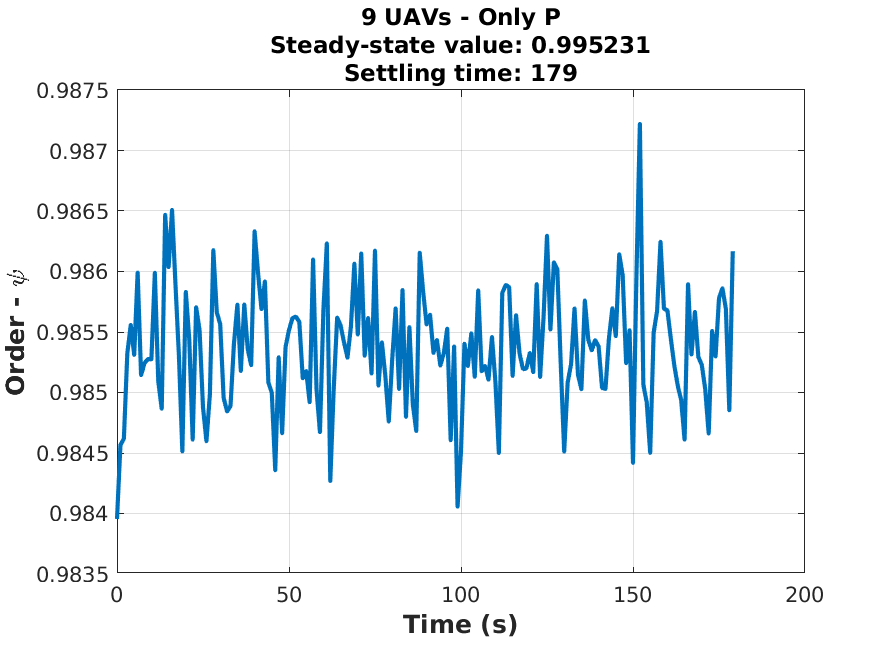}
  \caption{Order value in the experiment with real robots using the direct information exchange-based approach.}
  \label{fig:real_gps_order}
\end{figure}

During the experiment, the flocking moved in an arbitrary direction, as intended (see \reffig{fig:real_gps_heading}). After reaching the boundary of the workspace, the robots are triggered to finish the mission and land. The experiment revealed a steady-state value of $\Bar{\mu}=0.9952$, demonstrating the success of our approach, as it indicates that the order assumed values close to $1$ during the experiment (see \reffig{fig:real_gps_order}).

\subsubsection{Results with the UVDAR system}

\begin{figure}[t]
  \centering
  \includegraphics[width=0.5\textwidth]{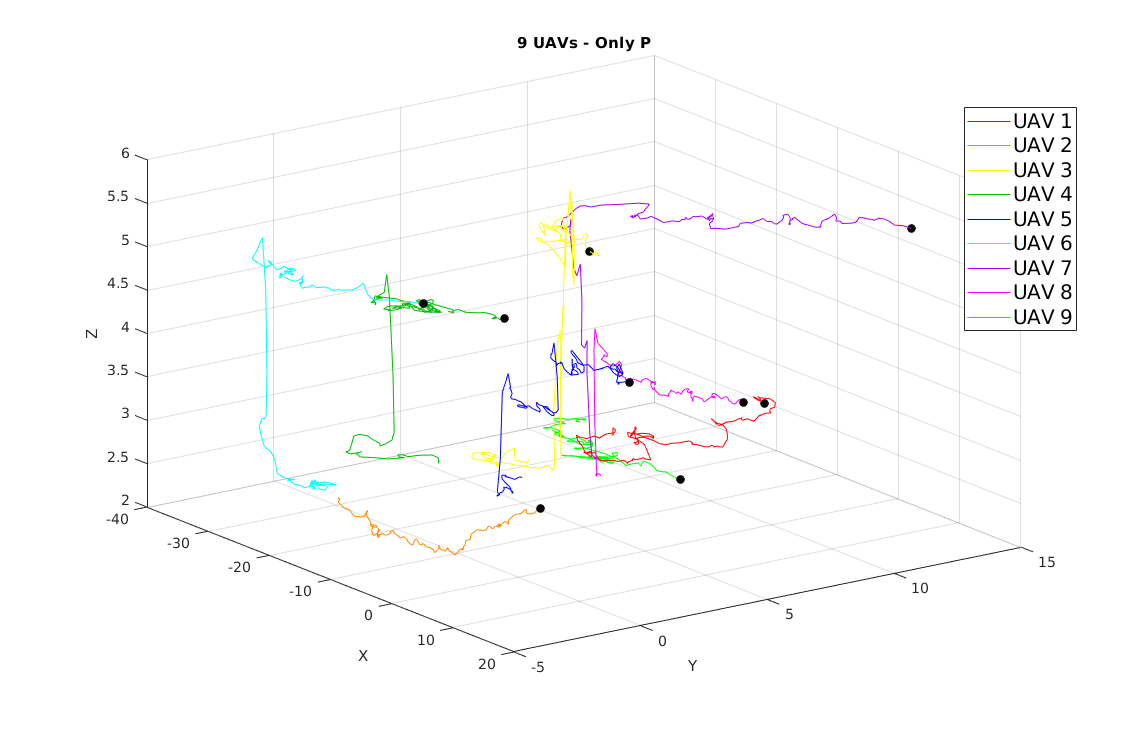}
  \caption{The 3D plot of the UAV flocking using the \ac{uvdar} system.}
  \label{fig:real_uvdar_blind_spot}
\end{figure}

In the experiment with real robots using the \ac{uvdar} as a visual relative localization approach, the swarming starts 10s after, and again, the UAVs rapidly converge into a cohesive flocking. We again used nine UAVs in the flocking, although the experiments were performed in another part of the desert. For a while, the robots converged again in range and bearing and moved in an arbitrary direction (see \reffig{fig:real_uvdar_blind_spot}), as intended.

\begin{figure}[t]
  \centering
  \includegraphics[width=0.5\textwidth]{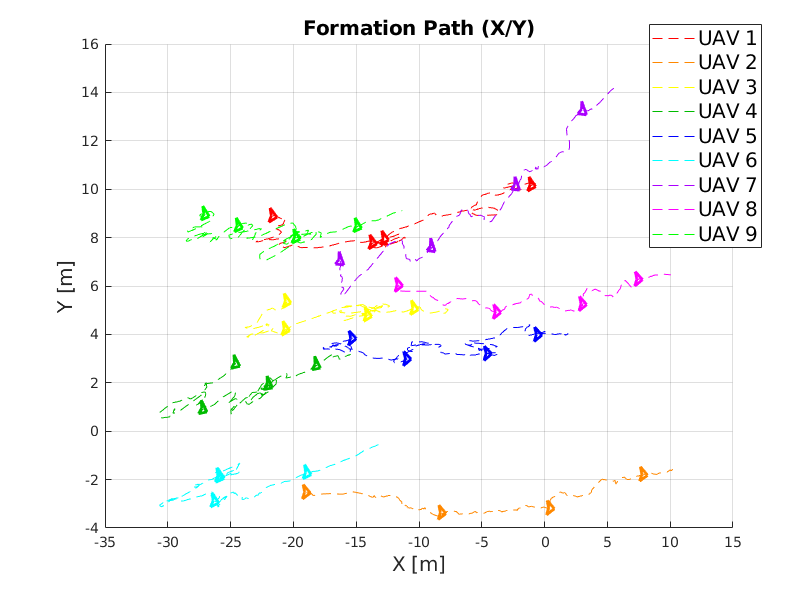}
  \caption{XY plot showing the heading of the flocking during the experiment with real robots using the \ac{uvdar} system.}
  \label{fig:real_uvdar_heading}
\end{figure}

\begin{figure}
  \centering
  \includegraphics[width=0.5\textwidth]{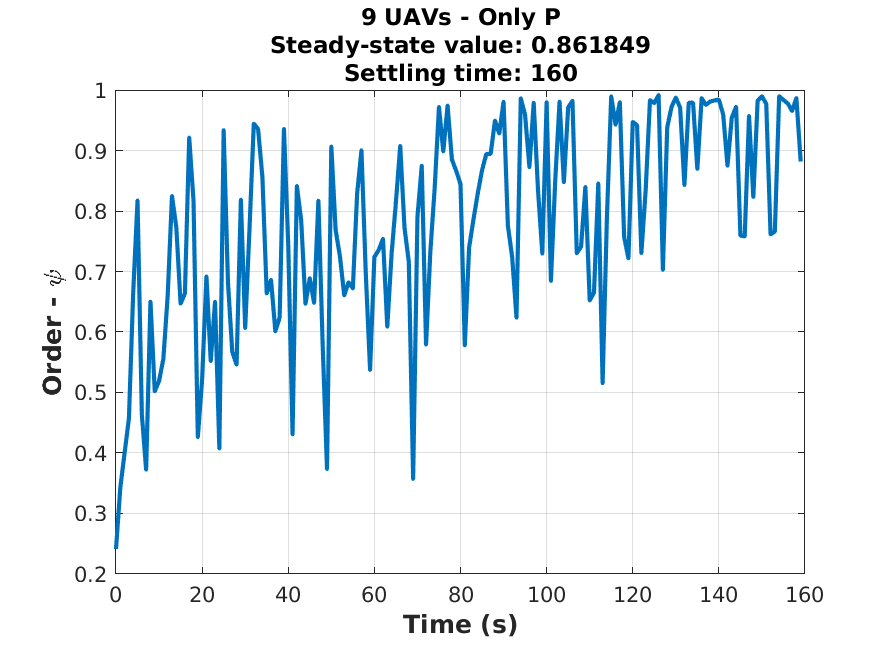}
  \caption{Order value in the experiment with real robots using the \ac{uvdar} system.}
  \label{fig:real_uvdar_order}
\end{figure}

This experiment revealed a steady-state value of $\Bar{\mu}=0.8619$. To better understand the not-so-high result of the steady-state value, we need to understand that the \ac{uvdar} is a camera-based relative localization system. It relies on visual estimation, which can lead to a higher covariance of observation. In this work, we only use the mean estimation. However, we already have considered the use of the covariance within the range and bearing extraction in future works. Furthermore, \reffig{fig:real_uvdar_heading} also shows that the UAVs tend to converge into subgroups, although these subgroups are, somehow close to each other. After a while, the \ac{uav} subgroups can no longer detect the other UAVs (due to the cut-off seen in \refasm{asm:flocking}). These sensory limitations can be seen in \reffig{fig:real_uvdar_order} by observing the variation of the order. Nevertheless, the proposed swarming approach was able to stabilize the swarm and achieve the required behavior.

\section{Conclusion}
\label{sec:conclusion}

In this work, we have presented a simple method for \ac{uav} flocking based only on a proximal function. This function has been used to enable the \acp{uav} not only to maintain a cohesive \ac{3D}-shape flock but to move in an arbitrary direction. We performed two sets of experiments, one using \ac{gps} and communication devices to exchange \ac{uav} positions and estimate the range and bearing, while the other set of experiments used only \ac{uvdar} to estimate the range and the bearing with no communication between \acp{uav}.

Figures \ref{fig:real_gps_order} and \ref{fig:real_uvdar_order} showed that our approach enabled the robots to converge towards the flock and maintain the cohesiveness of the group despite obstacles such as dunes. In addition, we achieved self-organized flocking with limited sensory information (either only GPS or only UVDAR) for aerial robots subjected to an environment with high dynamics such as wind (desert winds). Within the novel flocking approach itself, we have provided a framework and guidelines to enable fundamental achievements of swarm research to be integrated into \ac{uav} systems working in real-world conditions.

\section*{Declarations}

\subsection*{Acknowledgments}

We thank the \ac{mrs} hardware Eng. Daniel Hert and postdoc Tomas Baca for helping with the experiments.

\subsection*{Funding}
This work was supported by the Technology Innovation Institute — Sole Proprietorship LLC, UAE, under research project No. TII/ATM/2032/2020, by CTU grant no SGS23/177/OHK3/3T/13, by the Czech Science Foundation (GAČR) under research project No. 23-07517S, by the Europen Union under the project Robotics and advanced industrial production (reg. no. $CZ.02.01.01/00/22\_008/0004590$), by the National Council for Scientific and Technological Development – CNPq, by the National Fund for Scientific and Technological Development – FNDCT, and by the Ministry of Science, Technology and Innovations – MCTI from Brazil under research project No. 304551/2023-6 and 407334/2022-0, and by the Paraiba State Research Support Foundation - FAPESQ under research project No. 3030/2021.

\subsection*{Conflicts of interest/Competing interests}
Not applicable.

\subsection*{Code or data availability}
Not applicable.

\subsection*{Authors' Contributions}


All authors contributed to the Conceptualization. Thulio Amorim carried out Methodology, Software, Data curation, and Writing — Original draft. Tiago Nascimento performed Writing — Original draft, Supervision, and Visualization. Akash Chaudhary and Tomas Baca were involved in Investigation. Eliseo Ferrante and Martin Saska assisted with Funding acquisition, Resources, Verification, and Writing — Review \& Editing. All authors commented on the previous versions and approved the final manuscript.

\subsection*{Ethics approval}
Not applicable.

\subsection*{Consent to participate}
Not applicable.

\subsection*{Consent for publication}
Not applicable.

\bibliographystyle{spmpsci}
\bibliography{main}

\end{document}

%% file: tikz_config.tex
\tikzset{
  state/.style={
    rectangle,
    draw=black, very thick,
    minimum height=1.0em,
    text centered,
  },
  smallstate/.style={
    rectangle,
    draw=black, very thick,
    minimum height=0.2em,
    text centered,
  },
  final_state/.style={
    rectangle,
    rounded corners,
    draw=black, very thick,
    minimum height=2em,
    text centered,
  },
  initial_state/.style={
    rectangle,
    double=white,
    double distance=1pt,
    inner sep=2pt,
    draw=black, very thick,
    minimum height=2em,
    text centered,
  },
  color_state/.style={
    rectangle,
    draw=black, very thick,
    fill=yellow!40,
    minimum height=2em,
    text centered,
  },
  point/.style={
    circle,
    inner sep=0pt,
    minimum size=3pt,
    fill=red
  },
  adder/.style={
    circle,
    inner sep=2pt,
    minimum size=0.3in,
    draw=black, very thick,
    text centered
  },
  state_gray/.style={
    rectangle,
    draw=black, very thick,
    fill=gray!40,
    minimum height=2.0em,
    text centered,
  },
  state_white/.style={
    rectangle,
    draw=black, very thick,
    fill=white,
    minimum height=1.0em,
    text centered,
    text=black,
    inner sep=0,
  },
  state_green/.style={
    rectangle,
    draw=black, very thick,
    fill=green!50,
    minimum height=1.0em,
    text centered,
    text=black,
    inner sep=0,
  },
  state_red/.style={
    rectangle,
    draw=black, very thick,
    fill=red!70,
    minimum height=1.0em,
    text centered,
    text=black,
    inner sep=0,
  },
  state_org/.style={
    rectangle,
    draw=black, very thick,
    fill=orange!40,
    minimum height=1.0em,
    text centered,
    text=black,
    inner sep=0,
  },
  state_blue/.style={
    rectangle,
    draw=black, very thick,
    fill=blue!40,
    minimum height=1.0em,
    text centered,
    text=black,
    inner sep=0,
  },
  final_state/.style={
    rectangle,
    rounded corners,
    draw=black, very thick,
    minimum height=2em,
    text centered,
  },
  initial_state/.style={
    rectangle,
    double=white,
    double distance=1pt,
    inner sep=2pt,
    draw=black, very thick,
    minimum height=2em,
    text centered,
  },
  point/.style={
    circle,
    inner sep=0pt,
    minimum size=3pt,
    fill=red
  },
}

%% file: definitions.tex
\newcommand{\usvstateworld}{\mathbf{b_w}}
\newcommand{\usvstateworldoscillatory}{\mathbf{b_w^O}}
\newcommand{\usvinput}{\mathbf{u_b}}
\newcommand{\dragacc}{A_d}
\newcommand{\dragaccx}{a_{dx}}
\newcommand{\dragaccy}{a_{dy}}
\newcommand{\usvstatetransitionmat}{\mathbf{A_b}}
\newcommand{\usvstateinputmat}{\mathbf{B_b}}
\newcommand{\usvcovariance}{\mathbf{Q_b}}
\newcommand{\worldtobody}{^b\mathbf{R}_w}
\newcommand{\bodytoworld}{^w\mathbf{R}_b}
\newcommand{\coeffdragy}{Cd_y}
\newcommand{\droneinputvector}{\mathbf{u}}
\newcommand{\dronestatematrix}{\mathbf{D}}
\newcommand{\droneinputmatrix}{\mathbf{E}}
\newcommand{\dronestatevector}{\mathbf{x}}
\newcommand{\dronestatevectdes}{\overset{*}{\dronestatevector}}
\newcommand{\substatematrix}{\mathbf{D'}}
\newcommand{\subinputmatrix}{\mathbf{E'}}
\newcommand{\headingusv}{\eta}
\newcommand{\headinguav}{\psi}
\newcommand{\deltapred}{\Delta t_{p}}
\newcommand{\deltat}{dt}

\acrodef{uav}[UAV]{uncrewed aerial vehicle}
\acrodef{mav}[MAV]{multi-rotor aerial vehicle}
\acrodef{mpc}[MPC]{model predictive control}
\acrodef{nmpc}[NMPC]{nonlinear model predictive control}
\acrodef{usv}[USV]{uncrewed surface vehicle}
\acrodef{fft}[FFT]{fast Fourier transform}
\acrodef{ode}[ODE]{ordinary differential equation}
\acrodef{dof}[DOF]{degrees of freedom}
\acrodef{imu}[IMU]{inertial measurement unit}

\acrodef{2D}[2D]{two-dimensiona}
\acrodef{3D}[3D]{three-dimensional}
\acrodef{CACOC}[CACOC]{ant colony algorithm with chaotic dynamics}
\acrodef{fov}[FOV]{field of view}
\acrodef{gcs}[GCS]{ground control station}
\acrodef{gnss}[GNSS]{global navigation satellite system}
\acrodef{gps}[GPS]{global positioning system}
\acrodef{ipso}[IPSO]{improved particle swarm optimization}
\acrodef{led}[LED]{light-emitting diode}
\acrodef{mrs}[MRS]{multi-robot systems}
\acrodef{msos}[MSOS]{modified symbiotic organisms search}
\acrodef{SLERP}[SLERP]{spherical linear interpolation}
\acrodef{SPSO}[SPSO]{spherical vector-based particle swarm optimization}
\acrodef{TSS}[TSS]{time stamp segmentation}
\acrodef{ugv}[UGV]{unmanned ground vehicle}
\acrodef{UV}[UV]{ultraviolet}
\acrodef{uvdar}[UVDAR]{ultraviolet direction and ranging}

%% file: fig_system_archicteture.tex
\pgfdeclarelayer{foreground}
\pgfsetlayers{background,main,foreground}

\tikzset{radiation/.style={{decorate,decoration={expanding waves,angle=90,segment length=4pt}}}}

\begin{tikzpicture}[->,>=stealth', node distance=3.0cm,scale=1.0, every node/.style={scale=1.0}]


  \node[color_state, shift = {(0.0, 0.0)}] (navigation) {
      \begin{tabular}{c}
        \footnotesize Flocking\\
        \footnotesize Controller
      \end{tabular}
    };

\node[state_gray, right of = navigation, shift = {(0.2, 0)}] (tracker) {
      \begin{tabular}{c}
        \footnotesize MPC \\
        \footnotesize tracker
      \end{tabular}
    };

  \node[state_gray, right of = tracker, shift = {(0.1, 0)}] (controller) {
      \begin{tabular}{c}
        \footnotesize Position/Attitude \\
        \footnotesize controller
      \end{tabular}
    };
    
  \node [color_state, above of = controller, shift = {(-0.3, 0.2)}] (extractor) {
    \begin{tabular}{c}
      \small Range and Bearing \\
      \small Extractor
    \end{tabular}
  };

  \node[state, right of = controller, shift = {(0.4, -0)}] (attitude) {
      \begin{tabular}{c}
        \footnotesize Attitude rate\\
        \footnotesize controller
      \end{tabular}
    };
    
  \node[smallstate, below of = attitude, shift = {(-0.6, 2.1)}] (imu) {
      \footnotesize IMU
    };

  \node[state, right of = attitude, shift = {(0.0, -0)}] (actuators) {
      \begin{tabular}{c}
        \footnotesize UAV \\
        \footnotesize actuators
      \end{tabular}
    };
    
  \node[state_org, above of = actuators, shift = {(-1.8, -0.5)}] (uvdar) {
      \begin{tabular}{c}
        \footnotesize UVDAR\\
        \footnotesize System
      \end{tabular}
    };
    
 \node[state_org, above of = uvdar, shift = {(-0.0, -1.5)}] (comm) {
      \begin{tabular}{c}
        \footnotesize Intra-Swarm\\
        \footnotesize Communication
      \end{tabular}
    };

  \node[state, right of = actuators, shift = {(-1.1, -0)}] (sensors) {
      \begin{tabular}{c}
        \footnotesize Onboard \\
        \footnotesize sensors
      \end{tabular}
    };

  \node[state_gray, below of = attitude, shift = {(0, 0.9)}] (estimator) {
      \begin{tabular}{c}
        \footnotesize State \\
        \footnotesize estimator
      \end{tabular}
    };

  \node[state_gray, right of = estimator, shift = {(0.2, 0.0)}] (localization) {
      \begin{tabular}{c}
        \footnotesize Odometry \& \\
        \footnotesize localization
      \end{tabular}
    };



 \path[->] ($(navigation.east) + (0.0, 0)$) edge [] node[above, midway, shift = {(0.0, 0.05)}] {
      \begin{tabular}{c}
        \footnotesize $\mathbf{r}_d, \eta_d$\\
        \footnotesize \textit{on demand}
    \end{tabular}} ($(tracker.west) + (0.0, 0.00)$);

  \path[->] ($(tracker.east) + (0.0, 0)$) edge [] node[above, midway, shift = {(0.0, 0.05)}] {
      \begin{tabular}{c}
        \footnotesize $\bm{\chi}_d$\\
        \footnotesize \SI{100}{\hertz}
    \end{tabular}} ($(controller.west) + (0.0, 0.00)$);

  \path[->] ($(tracker.south |- estimator.west) + (0.0, 0.0)$) edge [dotted] node[left, midway, shift = {(0.2, 0.00)}] {
      \begin{tabular}{r}
        \scriptsize initialization\\[-0.5em]
        \scriptsize only
    \end{tabular}} ($(tracker.south) + (0.0, 0.00)$);

  \path[->] ($(controller.east) + (0.0, 0)$) edge [] node[above, midway, shift = {(0.0, 0.05)}] {
      \begin{tabular}{c}
        \footnotesize $\bm{\omega}_d$\\
        \footnotesize $T_d$ \\
        \footnotesize \SI{100}{\hertz}
    \end{tabular}} ($(attitude.west) + (0.0, 0.00)$);

  \draw[-] ($(controller.south)+(0.25,0)$) -- ($(controller.south |- estimator.west) + (0.25, 0.25)$) edge [->] node[above, near start, shift = {(-0.2, 0.05)}] {
      \begin{tabular}{c}
        \footnotesize $\mathbf{a}_d$
    \end{tabular}} ($(estimator.west) + (0, 0.25)$);

  \path[->] ($(attitude.east) + (0.0, 0)$) edge [] node[above, midway, shift = {(0.0, 0.05)}] {
      \begin{tabular}{c}
        \footnotesize $\bm{\tau}_d$ \\
        \footnotesize $\approx$\SI{1}{\kilo\hertz}
    \end{tabular}} ($(actuators.west) + (0.0, 0.00)$);

  \path[-] ($(estimator.west)+(0, 0)$) edge [] node[above, near start, shift = {(-1.0, 0.0)}] {
      \begin{tabular}{c}
        \footnotesize $\mathbf{x}$, $\mathbf{R}$, $\bm{\omega}$\\
        \footnotesize \SI{100}{\hertz}
    \end{tabular}} ($(navigation.south |- estimator.west)$) -- ($(navigation.south |- estimator.west)$) edge [->,] ($(navigation.south)+(0, 0)$);

  \path[->] ($(controller.south |- estimator.west)+(0, 0)$) edge [] ($(controller.south)$);

  \draw[-] ($(imu.east) + (0.0, 0.0)$) -- ($(estimator.north |- imu.east) + (0.3, 0)$) edge [->] node[right, midway, shift = {(-0.2, 0.3)}] {
      \begin{tabular}{c}
        \footnotesize $\mathbf{R}$, $\bm{\omega}$
    \end{tabular}} ($(estimator.north) + (0.3, 0.0)$);

  \draw[-] ($(sensors.south)+(0, 0)$) -- ($(sensors.south |- localization.east)$) edge [->] ($(localization.east)$);
  \draw[-] ($(sensors.south)+(0.1, 0)$) -- ($(sensors.south |- localization.east) + (0.1, -0.1)$) edge [->] ($(localization.east) + (0.0, -0.1)$);
  \draw[-] ($(sensors.south)+(-0.1, 0)$) -- ($(sensors.south |- localization.east) + (-0.1, 0.1)$) edge [->]  ($(localization.east) + (0.0, 0.1)$);

  \draw[->] ($(localization.west)+(0, 0)$) -- ($(estimator.east)$);
  \draw[->] ($(localization.west)+(0, 0.1)$) -- ($(estimator.east) + (0, 0.1)$);
  \draw[->] ($(localization.west)+(0, -0.1)$) -- ($(estimator.east) + (0, -0.1)$);

  \draw[-] ($(sensors.north) + (0, 0)$) -- ($(sensors.north |- uvdar.east)$) edge [->]
  ($(uvdar.east)+(0, 0)$);

  \draw[-] ($(sensors.north) + (0, 0)$) -- ($(sensors.north |- comm.east)$) edge [->]
  ($(comm.east)+(0, 0)$);

  \draw[-] ($(extractor.west) + (0, 0)$) -- ($(extractor.west -| navigation.north)$) edge [->] 
  node[above, midway, shift = {(1.5, 1.2)}] {
      \begin{tabular}{c}
        \footnotesize UAV's teammate $d_n$, $\phi_n$, $\theta_n$
    \end{tabular}} ($(navigation.north)+(0, 0)$);

  \draw[-] ($(comm.west) + (0, 0)$)[dotted] -- ($(comm.west -| extractor.north)$) edge [dotted] [->] 
  node[above, midway, shift = {(2, 0)}] {
      \begin{tabular}{c}
        \footnotesize UAV's teammate Global position
    \end{tabular}} ($(extractor.north)+(0, 0)$);
  
  \draw[-] ($(uvdar.west) + (0, 0)$)[dotted] -- ($(uvdar.west -| extractor.south)$) edge [dotted] [->] 
  node[below, midway, shift = {(2, 0)}] {
      \begin{tabular}{c}
        \footnotesize UAV's teammate Relative position
    \end{tabular}} ($(extractor.south)+(0, 0)$);

  \begin{pgfonlayer}{background}
    \path (attitude.west |- attitude.north)+(-0.45,0.8) node (a) {};
    \path (imu.south -| sensors.east)+(+0.25,-0.20) node (b) {};
    \path[fill=gray!3,rounded corners, draw=black!70, densely dotted]
      (a) rectangle (b);
  \end{pgfonlayer}
  \node [rectangle, above of=actuators, shift={(-0.6,0.55)}, node distance=1.7em] (autopilot) {\footnotesize UAV plant};

  \begin{pgfonlayer}{background}
    \path (attitude.west |- attitude.north)+(-0.25,0.47) node (a) {};
    \path (imu.south -| attitude.east)+(+0.25,-0.10) node (b) {};
    \path[fill=gray!3,rounded corners, draw=black!70, densely dotted]
      (a) rectangle (b);
  \end{pgfonlayer}
  \node [rectangle, above of=attitude, shift={(0,0.2)}, node distance=1.7em] (autopilot) {\footnotesize Pixhawk autopilot};


\end{tikzpicture}